\documentclass{article}
\usepackage{PRIMEarxiv}
\usepackage[utf8]{inputenc}
\usepackage[T1]{fontenc}
\usepackage{amsmath,amssymb,booktabs,array,graphicx,microtype}
\usepackage{xcolor,hyperref,url,fancyhdr,placeins}
\hypersetup{colorlinks=true,linkcolor=blue!55!black,citecolor=blue!55!black,urlcolor=teal!65!black,
 pdftitle={Causal Latent Workspaces for Looped Transformers},pdfauthor={Ilia Zaznov}}
\graphicspath{{figures/}}
\newcommand{\LN}{\operatorname{LN}}
\newcommand{\softmax}{\operatorname{softmax}}
\newcommand{\sigmoid}{\operatorname{sigmoid}}
\newcommand{\R}{\mathbb{R}}
\newcommand{\code}[1]{\texttt{\detokenize{#1}}}
\title{Causal Latent Workspaces for Looped Transformers}
\author{Ilia Zaznov\\\texttt{izaznov@gmail.com}}
\date{October 2026}
\begin{document}
\maketitle
\begin{abstract}
Reusing a Transformer block across depth increases computation without
replicating its parameters, but successive passes have no dedicated state
through which positions can exchange compact summaries of their causal
prefix. We introduce the \emph{causal latent workspace Transformer}: an
anchored, four-round shared-depth architecture in which each position
writes gated vectors into prefix-specific latent slots and retrieves a
content-addressed summary before the next shared block application. This
path complements causal self-attention while the original input embedding
remains accessible at every round. Across three byte-level corpora,
three seeds, and two matched parameter budgets, the proposed model
achieves the best mean held-out perplexity in five of six settings
against five Transformer controls. At 101.71M parameters it leads on
WikiText-103 raw, enwik8, and TinyStories, reaching perplexities
2.7820, 2.7843, and 1.6823; these improve upon the strongest
non-workspace control on each corpus by 1.06\%, 0.73\%, and 1.01\%.
All nine seed-labelled comparisons at this budget favor the workspace.
At 0.918M parameters it leads on WikiText-103 and TinyStories, whereas
a conditioned loop is stronger on enwik8. Architectural ablations show
that adding the workspace improves upon the anchored loop in all six
settings and support refreshing its causal summaries at each round.
Together, these results identify a causal,
content-addressable prefix workspace as an effective structural
augmentation of looped Transformers.
\end{abstract}

\section{Introduction}
Self-attention provides the Transformer with a direct, parallel mechanism
for integrating information across a sequence \cite{vaswani}. Its
effectiveness also raises a continuing architectural question: how should
this information be organized as representations are refined through
depth? In a conventional stack, independently parameterized layers
successively transform the hidden sequence. A looped Transformer instead
reuses a shared block, allocating more computation to refinement without
replicating the block's parameters \cite{universal,geiping}. This makes
the transition between rounds a central design choice. A standard shared
loop carries forward the hidden sequence, but provides no separate,
addressable representation of the causal prefix.

We introduce the \emph{causal latent workspace Transformer}, which augments
shared-depth refinement with a learned write/read pathway. At each round,
gated writes accumulate information from the causal prefix into a small
set of vector slots. A query derived from the current token state retrieves
a weighted summary from these slots before the next shared block
application. The original input embedding remains available through an
anchor, allowing refinement to combine the input, the evolving hidden
sequence, and a content-addressed prefix summary. Causal self-attention
and the workspace consequently provide complementary routes: the former
operates on individual token states, while the latter operates on learned
prefix summaries. Each prefix has its own workspace, reconstructed from
the current round and reset at the next input block.

Our experiments examine whether this additional information pathway
improves language modeling beyond shared depth, input anchoring, and
more expressive recurrent transitions. We compare the proposed model
with five Transformer controls at approximately 0.918M and 101.71M
parameters, using the same training-target budget within each size.
Across WikiText-103 raw, enwik8, and TinyStories, with three seeds per
condition, the workspace achieves the lowest mean held-out perplexity
in five of six corpus--size settings. It improves upon the anchored
loop in all six settings. At 101.71M parameters it also outperforms
the strongest alternative on every corpus and for every corresponding
seed label. At the smaller budget it leads on WikiText-103 and
TinyStories, while the conditioned-dense control leads on enwik8.
These comparisons establish the value of a causal workspace under
matched parameter and training-target budgets.

The contribution is a causal write/read mechanism that makes compact
prefix summaries available throughout recurrent refinement, together
with an empirical study of its effect relative to alternative recurrent
architectures. The following sections place the design in the context
of recurrent depth and neural memory, derive the architecture, and
illustrate its computation with a fully specified example. We then
describe the matched evaluation and use architectural ablations to
distinguish the benefit of adding a workspace from the choice of how
its state is updated across rounds.

\section{Related work}
Recurrent depth provides a starting point for organizing repeated
computation. Universal Transformers share a transition over depth
\cite{universal}, while adaptive computation time varies the number of
applications according to the input \cite{act}. Block-Recurrent
Transformers combine attention with recurrent block-level state
\cite{blockrecurrent}. More recent latent-recurrence and
Mixture-of-Recursions architectures study shared computation and selective
depth allocation \cite{geiping,mor}. Together, these approaches establish
repeated refinement as an alternative to an independently parameterized
stack. The present work addresses the transition within that refinement:
we keep the number of shared-block applications fixed and introduce a
separate pathway for constructing and retrieving causal prefix summaries.

Neural memory provides the second foundation of the design.
Transformer-XL carries representations across segments \cite{txl},
Compressive Transformers compress older states \cite{compressive},
and Memorizing Transformers retrieve past key--value pairs
\cite{memorizing}. Differentiable memory systems such as Neural Turing
Machines and Differentiable Neural Computers separate a controller from
an addressable store \cite{ntm,dnc}. These approaches extend accessible
history or maintain a store as a sequence is processed. The causal
workspace instead organizes information within the current context
during depth refinement. It reconstructs its summaries from the current
token states at every round, allowing the write/read pathway to respond
to changes in those states. This function is distinct from a decoder
key--value cache, which reuses previously computed attention projections
during autoregressive decoding.

Latent bottlenecks offer a way to implement such a compact information
pathway. The shared global workspace of Goyal et al.\ coordinates
specialized modules \cite{goyal}; Slot Attention learns reusable latent
slots \cite{slots}; and the Perceiver exchanges information through a
compact latent array \cite{perceiver}. Applying this principle to causal
language modeling requires a separate summary at each sequence position.
A slot available at position $i$ can include writes only from positions
$j\leq i$. Our gated prefix accumulations satisfy this constraint while
sharing slot keys and projection parameters across positions and rounds.
The resulting workspace is a collection of causal summaries rather than
a single global summary of the entire sequence.

This separation of shared parameters from transient input-dependent state
also connects the model to fast-weight networks \cite{fastweights,schlag}.
In our architecture, the transient state is a gated vector accumulation;
the learned projection matrices remain fixed during a forward pass.
Other sequence architectures pursue different interventions:
Linear Transformers and Performers reformulate attention
\cite{linear,performer}, Longformer and Big Bird restrict interaction
patterns \cite{longformer,bigbird}, and Reformer changes attention and
memory organization \cite{reformer}. Mamba, RetNet, and Hyena use recurrent
or convolutional sequence mechanisms \cite{mamba,retnet,hyena}, while
FlashAttention improves the execution of exact attention \cite{flash}.
The workspace complements these directions by retaining full causal
self-attention and changing the information supplied between shared-depth
applications. Its effect is therefore evaluated through both prediction
quality and the additional computation required by the write/read path.

The experimental design follows from this architectural question.
WikiText-103 \cite{merity}, enwik8 \cite{hutter}, and TinyStories
\cite{tinystories} provide different text distributions on which to assess
the same mechanism. Since stochastic training can affect architectural
rankings \cite{dodge}, we evaluate the complete corpus--size grid with
three seeds and score final checkpoints over all complete held-out blocks.

\section{Architecture}
\label{sec:architecture}
The proposed model augments one shared causal Transformer block with
two complementary paths, shown in Figure~\ref{fig:architecture}.
The anchor preserves the original input representation as hidden states
are refined; the workspace gathers and redistributes prefix information
between successive applications of the block. We first define
next-byte prediction, then the write/read operations, and finally
their composition with shared-depth refinement.
\subsection{Prediction task and notation}
Let $\mathcal V=\{0,\ldots,256\}$ be the 257-symbol output vocabulary.
Symbols $0$--$255$ are byte values; symbol $256$ is the row separator used
for WikiText only. For a block of $L+1$ symbols
$(x_1,\ldots,x_{L+1})$, the model consumes $(x_1,\ldots,x_L)$ and predicts
$y_i=x_{i+1}$ at position $i$, where $1\leq i\leq L$.
We use $L=512$. Batch indices are suppressed in equations; sequences
in a batch are processed independently.

Let $d$ be the token-state width, $R$ the number of shared rounds,
$S$ the workspace slot count, and $m$ the coordinates per slot.
Experiments use $R=4$ and $S=4$. The trainable embedding lookup
$E:\mathcal V\rightarrow\R^d$ and trainable position vector $p_i\in\R^d$
produce anchor $e_i$. Superscript $r$ denotes the state before shared
round $r$, with $r=0,\ldots,R-1$:
\begin{equation}
 e_i=E(x_i)+p_i,\qquad h_i^0=e_i .
 \label{eq:embedding}
\end{equation}
Vectors are column vectors. Brackets $[a;b]$ concatenate vectors.
$\theta$ denotes all slowly learned parameters. Affine projections include
their displayed biases; weights and biases are shared across rounds unless
explicitly indexed by $r$.

\subsection{Anchoring and gated causal writes}
Each round first fuses the unchanged anchor with the current token state:
\begin{equation}
 u_i^r=\LN_a\!\left(h_i^r+F_a([e_i;h_i^r])\right),\qquad
 F_a(w)=A_2\,\mathrm{GELU}(A_1w+a_1)+a_2 .
 \label{eq:anchor}
\end{equation}
Here $w\in\R^{2d}$ is the concatenated argument,
$A_1\in\R^{d\times2d}$, $A_2\in\R^{d\times d}$,
and $a_1,a_2\in\R^d$ are learned parameters.
GELU is the element-wise activation of Hendrycks and Gimpel \cite{gelu}.
$\LN_a$ is a learned LayerNorm \cite{layernorm}; subscripts identify
separate parameter sets, not different mathematical operations.

With write-normalized state $z_i^r=\LN_w(u_i^r)$, the write gate
$g_{is}^r\in(0,1)$ and write vector $v_{is}^r\in(-1,1)^m$ for slot $s$ are
\begin{equation}
 g_{is}^r=\sigmoid((W_g z_i^r+b_g)_s),\qquad
 v_{is}^r=\tanh((W_v z_i^r+b_v)_{s,:}).
 \label{eq:write}
\end{equation}
$W_g\in\R^{S\times d}$, $b_g\in\R^S$,
$W_v\in\R^{Sm\times d}$, and $b_v\in\R^{Sm}$ are learned.
The value projection is reshaped into $S$ rows of length $m$;
$(s,:)$ selects a row. Sigmoid and tanh act element-wise.
The newly written memory for prefix $1,\ldots,i$ is
\begin{equation}
 N_{is}^r=\sum_{j=1}^{i}g_{js}^r v_{js}^r,\qquad
 D_{is}^r=\sum_{j=1}^{i}g_{js}^r,\qquad
 \widetilde M_{is}^r=\frac{N_{is}^r}{\sqrt{\max(D_{is}^r,\epsilon)}} ,
 \quad \epsilon=10^{-6}.
 \label{eq:prefix}
\end{equation}
$j$ indexes causal sources, $N_{is}^r\in\R^m$ is their gated vector
sum, $D_{is}^r$ their scalar gate mass, and
$\widetilde M_{is}^r\in\R^m$ the slot memory.
The current input position is included, but its next-symbol target is not.
The square-root denominator scales the accumulated value by gate mass
while preserving its dependence on the amount of accumulated evidence.
Unlike a weighted average, this normalization allows aligned writes
to increase the summary magnitude as the square root of prefix mass.

The main model recomputes $M_{is}^r=\widetilde M_{is}^r$ every round.
This choice lets each round revise its summary after the preceding
Transformer update. In a separate retention experiment, we replace
the direct assignment with
\begin{equation}
 M_{is}^0=\widetilde M_{is}^0,\qquad
 M_{is}^r=\rho_r M_{is}^{r-1}+(1-\rho_r)\widetilde M_{is}^r
 \ (r\geq1),\qquad \rho_r=\sigmoid(\ell_r).
 \label{eq:retention}
\end{equation}
Here $\ell_r$ is a learned scalar that controls the contribution of
the preceding round; its initialization $\log(0.2/0.8)$ gives
$\rho_r=0.2$. This variant blends old and newly written summaries
at the same prefix position. Both update rules reset workspace state
at input-block boundaries. Section~\ref{sec:results} compares their
effect under the same parameter and training-target budgets.

\subsection{Slot read and shared refinement}
Each slot has a learned key $k_s\in\R^m$, initialized with independent
normal coordinates of standard deviation $m^{-1/2}$.
The read query and attention over slots are
\begin{equation}
 q_i^r=W_q\LN_q(u_i^r)+b_q,\qquad
 \alpha_{is}^r=
 \frac{\exp((q_i^r)^\top k_s/\sqrt m)}
 {\sum_{t=1}^{S}\exp((q_i^r)^\top k_t/\sqrt m)} .
 \label{eq:read}
\end{equation}
$W_q\in\R^{m\times d}$ and $b_q\in\R^m$ are learned;
$t$ is a dummy slot index. Weights $\alpha_{is}^r$ are nonnegative
and sum to one over $s$.
The workspace residual $c_i^r\in\R^d$ and next hidden state are
\begin{equation}
 c_i^r=\gamma\!\left(W_o\sum_{s=1}^{S}\alpha_{is}^r M_{is}^r+b_o\right),
 \qquad h^{r+1}=T_\theta\!\left(\LN_b(u^r+c^r)\right).
 \label{eq:update}
\end{equation}
$W_o\in\R^{d\times m}$ and $b_o\in\R^d$ are learned;
$\gamma$ is a learned scalar initialized to $0.5$.
Unsubscripted $h^{r+1},u^r,c^r$ denote complete length-$L$ sequences.
$T_\theta$ is one pre-normalized causal Transformer block, reused
in all four rounds; its full equations appear in Appendix~\ref{app:block}.

\begin{figure}[t]
 \centering\includegraphics[width=\linewidth]{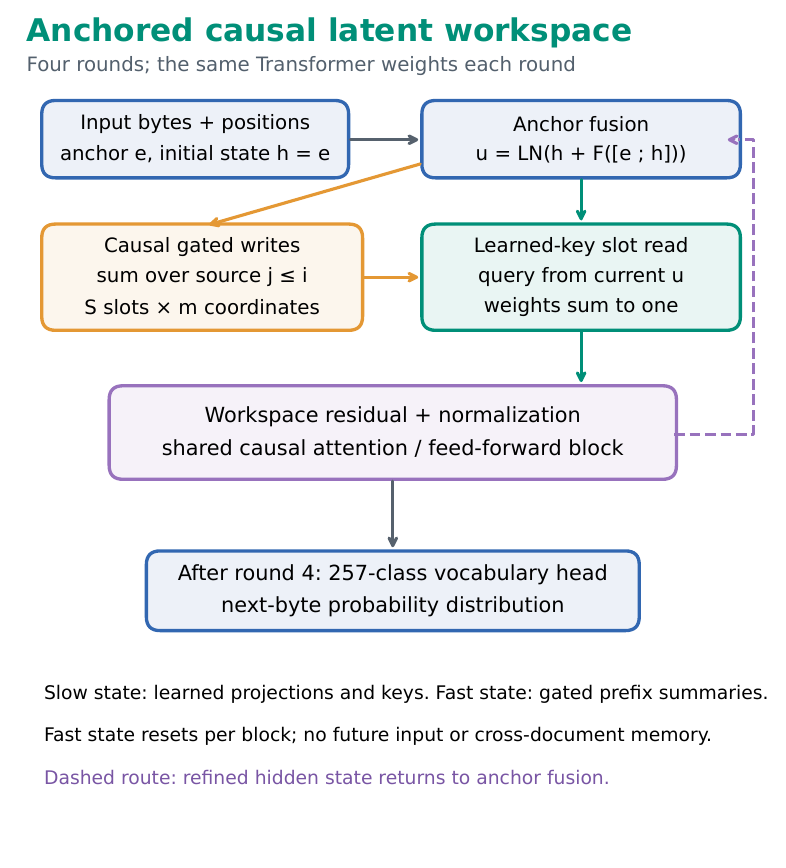}
 \caption{The evaluated architecture. Blue: input anchor and output head.
 Orange: gated causal prefix accumulation. Green: learned-key slot read.
 Purple: shared refinement and the recurrent route. Fast workspace
 activations are recomputed each round in the main model.}
 \label{fig:architecture}
\end{figure}

After the last round, the separately learned, bias-free vocabulary matrix
$W_{\mathrm{LM}}\in\R^{257\times d}$ produces logits $o_i$:
\begin{equation}
 o_i=W_{\mathrm{LM}}\LN_f(h_i^R),\qquad
 P_\theta(y_i=a\mid x_{1:i})=
 \frac{\exp(o_{i,a})}{\sum_{b\in\mathcal V}\exp(o_{i,b})}.
 \label{eq:head}
\end{equation}
$a$ is a candidate target, $b$ a dummy vocabulary index, and
$x_{1:i}$ the consumed prefix. Embedding and head weights are not tied.
Figure~\ref{fig:architecture} summarizes the resulting computation.
The input anchor and current state determine the causal writes and
the slot query. Their retrieved summary enters the shared block as
a residual, after which causal self-attention refines interactions
between token states. Repeating this sequence allows each round to
construct its workspace from the representations produced by the
preceding round.

\subsection{Causality and computational structure}
The workspace preserves the autoregressive structure of the shared block.
Initially $h_i^0$ depends only on $x_i$ and its position. Assume that
$h_j^r$ depends only on $x_{1:j}$. Anchor fusion and slot reads operate
position-wise, prefix accumulation at $i$ includes only $j\leq i$, and
causal attention masks $j>i$. Therefore $h_i^{r+1}$ depends only on
$x_{1:i}$. Retention combines states at the same prefix position and
preserves this property. By induction, each output depends only on its
consumed prefix.

The same construction permits all prefix summaries within a round to
be computed by causal cumulative sums. Slot keys and projection matrices
are shared learned parameters, whereas the summaries are input-dependent
activations. As the hidden sequence changes, each round recomputes its
writes and reads using those same parameters. For a batch of $B$ sequences,
parallel execution materializes a distinct $S\times m$ workspace at each
of the $L$ positions, giving an activation tensor of shape
$B\times L\times S\times m$.

For one sequence, the write projection requires work of order $LSmd$,
and prefix accumulation requires $LSm$. Query construction, slot
addressing, and the output projection add terms of order $Ldm$, $LSm$,
and $Lmd$, respectively. With feed-forward width $f$, the shared
Transformer block requires work of order $Ld^2+L^2d+Ldf$ per round.
Workspace operations are linear in context length for fixed $S,m,d$,
while the retained self-attention term is quadratic. Reusing parameters
across $R$ rounds thus changes the parameter allocation while executing
the block and workspace operations $R$ times.
Section~\ref{sec:results} examines the observed training cost alongside
prediction quality.

\FloatBarrier

\section{A byte-to-prediction walkthrough}
\label{sec:walkthrough}
\begin{figure}[t]
 \centering\includegraphics[width=\linewidth]{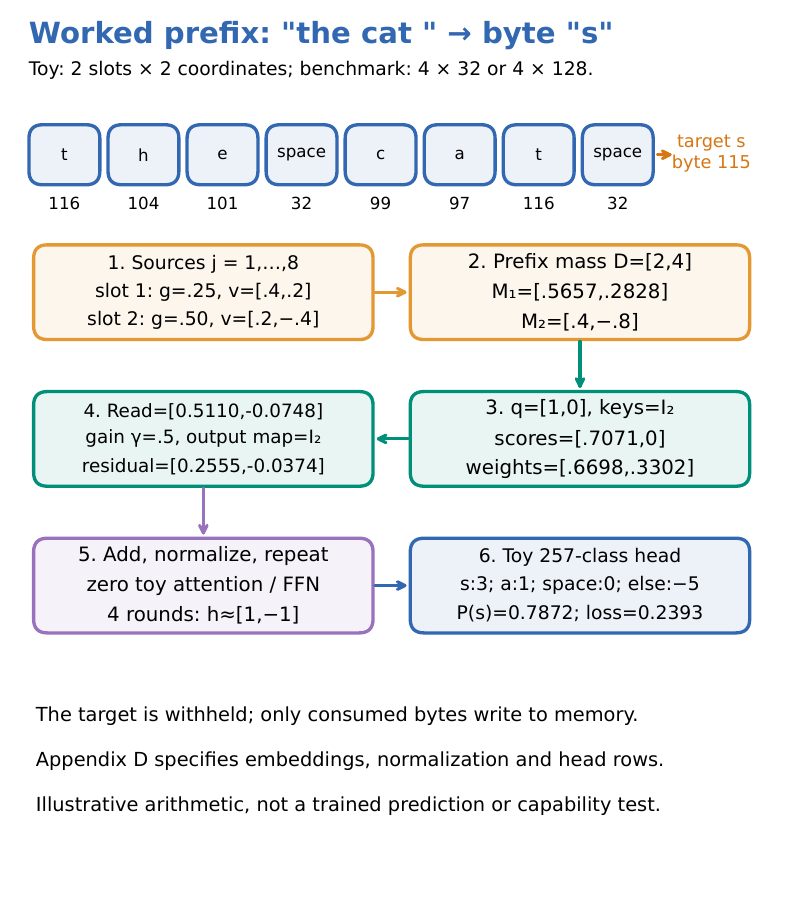}
 \caption{Numerical walkthrough: byte IDs, writes, prefix memories, slot
 addressing, residual, shared rounds, and a 257-class toy head.
 Appendix~\ref{app:toy} fully specifies the toy parameters.}
 \label{fig:walkthrough}
\end{figure}

The interaction between causal writes and slot reads can be illustrated
with the eight-byte prefix \texttt{the cat } and next-byte target
\texttt{s}. The input IDs are $(116,104,101,32,99,97,116,32)$, including
the two spaces; the target byte has ID $115$. Figure~\ref{fig:walkthrough}
traces this prefix through embedding, accumulation, addressing, recurrent
refinement, and the output distribution. To make the intermediate values
explicit, the example uses two slots with two coordinates each and a
fully specified set of parameters (Appendix~\ref{app:toy}). These
parameters define an arithmetic illustration of the architecture.

\begin{enumerate}
\item \textbf{Embed and anchor.} Equation~\ref{eq:embedding} turns each
integer ID into a vector and adds a position vector. In every round,
Equation~\ref{eq:anchor} exposes both the unchanged anchor and the
current refined state to the fusion network.
\item \textbf{Write from the consumed prefix.} At position $i=8$,
Equation~\ref{eq:prefix} uses all eight inputs. For a toy with
$S=m=2$, let every source have gates $(0.25,0.50)$ and slot vectors
$(0.4,0.2)$ and $(0.2,-0.4)$. Masses are $D_{8,1}=2$ and
$D_{8,2}=4$; numerators are $(0.8,0.4)$ and $(0.8,-1.6)$.
Dividing by square-root mass gives $M_{8,1}=(0.5657,0.2828)$
and $M_{8,2}=(0.4,-0.8)$. The round index is omitted because
these particular toy writes are constant.
\item \textbf{Address slots.} With query $q=(1,0)$ and keys
$k_1=(1,0)$, $k_2=(0,1)$, Equation~\ref{eq:read} gives scores
$(1/\sqrt2,0)$ and weights $(0.6698,0.3302)$.
This is attention over prefix summaries, not over individual tokens.
\item \textbf{Inject the read.} The weighted vector is
$(0.5110,-0.0748)$. With identity output map, zero bias,
and $\gamma=0.5$, the residual is $(0.2555,-0.0374)$.
The actual model learns the output map and gain, normalizes the sum,
and applies full causal attention plus a feed-forward network.
\item \textbf{Refine four times.} The same transformation is reused,
but the real model normally writes different summaries from updated
states. In the simple toy, zero attention/feed-forward weights
isolate workspace arithmetic; position eight has hidden state
approximately $(1,-1)$ after each round.
\item \textbf{Predict and score.} A specified toy head gives logit $3$
to \texttt{s}, $1$ to \texttt{a}, $0$ to space, and $-5$ to each
remaining symbol. The full 257-way softmax gives
$P(\texttt{s})=e^3/(e^3+e^1+1+254e^{-5})=0.7872$.
The single-target loss is $-\log(0.7872)=0.2393$ nats.
Training averages these losses over targets and updates slow parameters
by backpropagation. A target is never written before being predicted.
\end{enumerate}
\FloatBarrier

\section{Experimental protocol}
\label{sec:protocol}
We evaluate whether the causal workspace improves next-byte prediction
under a fixed parameter and training-target budget. Five controls
separate its contribution from that of shared depth, input anchoring,
and recurrent transition design. Within each model size, all conditions
use the same corpus representation, context length, optimizer recipe,
and number of scored training targets. The corpus and training
specifications are given in Tables~\ref{tab:data} and \ref{tab:recipe}.

\subsection{Five controls and parameter matching}
The six rows use one training/evaluation harness:
\begin{enumerate}
\item \textbf{Plain Transformer:} four independently parameterized
causal blocks and a final normalized vocabulary head.
\item \textbf{Fixed shared loop:} one block applied four times,
$h^{r+1}=T_\theta(h^r)$, without anchoring or workspace.
\item \textbf{Adaptive-exit mixture:} four shared passes and a
per-position mixture of their exits. An affine router on $e_i$
supplies four softmax weights; every exit is computed.
\item \textbf{Anchored shared loop:} Equation~\ref{eq:anchor} followed
directly by the shared block, without a workspace residual.
\item \textbf{Anchored conditioned-dense:} anchoring, learned
per-round element-wise scale/shift before the block, and a learned
readout from earlier exits. Additions initialize at zero, preserving
the anchored control's initial transition/readout.
\item \textbf{Causal workspace:} Equations~\ref{eq:embedding}--\ref{eq:head},
recomputing each round, without retention between rounds.
\end{enumerate}
Hidden widths and feed-forward dimensions are chosen to match
approximately 918K or 101.71M parameters, as detailed in
Appendix~\ref{app:hyper}. Every model executes four Transformer-block
applications. Relative to the workspace model, the maximum parameter
difference is 474 at the small budget and 6,058 at the larger budget.
Independent and shared-depth architectures allocate these parameters
differently: the larger plain Transformer has width 1,536, whereas the
fixed shared loop has width 3,072. The anchored and workspace models
both use width 2,560 and 20 heads; adding the workspace is compensated
by a narrower feed-forward sublayer. The comparison therefore measures
the effectiveness of each complete architecture at a common parameter
budget. The controls are implemented within the same experimental
framework, with their equations and dimensions specified in the appendices.

\subsection{Corpora and preprocessing}
The evaluation spans three complementary text distributions.
WikiText-103 raw contains Wikipedia articles and surrounding text,
enwik8 provides a continuous Wikipedia-derived byte stream, and
TinyStories contains short synthetic narratives. Their corpus
descriptions are given by Merity et al.\ \cite{merity}, the Hutter
Prize specification \cite{hutter}, and Eldan and Li \cite{tinystories}.
Table~\ref{tab:data} specifies the training pools, held-out sources,
and scored counts. Evaluation examples are excluded from gradient updates.

\begin{table}[t]
 \centering\small
 \caption{Data protocol. Held-out evaluation scores every complete
 513-symbol block from the designated evaluation stream.
 All models retain a 257-class head.}
 \label{tab:data}
 \begin{tabular}{p{1.03in}p{2.58in}p{2.37in}}
 \toprule
 Corpus & Training source and representation & Held-out evaluation stream \\
 \midrule
 WikiText-103 raw & Official \code{wikitext-103-raw-v1} training
 Parquet shards; UTF-8 bytes; blank rows skipped; symbol 256 appended
 to each nonempty row. & Official held-out split: 2,235 complete
 blocks, 1,144,320 scored targets. \\
 enwik8 & Official 100,000,000-byte file; first 90,000,000 bytes
 form the training pool. Raw bytes; no inserted separator. & Next
 5,000,000 bytes: 9,746 blocks, 4,989,952 scored targets.
 Final 5,000,000 bytes reserved and not reported. \\
 TinyStories & \code{TinyStoriesV2-GPT4-train.txt}; first
 250,000,000 bytes form the training pool. Raw file bytes,
 without inserted separators or story-wise parsing. & Entire
 \code{TinyStoriesV2-GPT4-valid.txt}: 43,864 blocks,
 22,458,368 scored targets. \\
 \bottomrule
 \end{tabular}
\end{table}
The source datasets are available from
\href{https://huggingface.co/datasets/Salesforce/wikitext}{Salesforce/WikiText},
\href{https://mattmahoney.net/dc/enwik8.zip}{enwik8}, and
\href{https://huggingface.co/datasets/roneneldan/TinyStories}{TinyStories}.
WikiText text rows are encoded as UTF-8 bytes, with symbol 256 appended
after each nonempty row. enwik8 and TinyStories retain their raw file
bytes without inserted separators. A common 257-class vocabulary is
used for all conditions, keeping the output head identical in form
across corpus representations.

Each stream is partitioned into non-overlapping blocks of 513 symbols.
A block supplies 512 input positions and their shifted next-symbol
targets; its first symbol provides context only, and any incomplete
trailing block is discarded. Hidden states, workspace summaries, and
positions reset at each block boundary. All architectures consequently
use the same within-block context, including when a boundary falls
inside an article or story. This common segmentation makes the scored
training and evaluation targets directly comparable within each corpus.

\subsection{Training, seeds, and evaluation}
Table~\ref{tab:recipe} specifies the two size-dependent recipes.
Within a size, the architecture changes but the total number of
scored training targets does not.
\begin{table}[t]
 \centering\small
 \caption{Training recipe within each parameter budget. Microbatch steps
 count forward/backward calls; optimizer updates account for accumulation.}
 \label{tab:recipe}
 \begin{tabular}{lrr}
 \toprule
 Setting & Small (0.918M) & Larger (101.71M) \\
 \midrule
 Context / output vocabulary & 512 / 257 & 512 / 257 \\
 Scored targets per run & 81,920,000 & 81,920,000 \\
 Microbatch / accumulation & 8 / 1 & 4 / 2 \\
 Microbatch steps / updates & 20,000 / 20,000 & 40,000 / 20,000 \\
 Effective sequences per update & 8 & 8 \\
 Learning rate & $3\times10^{-4}$ & $10^{-4}$ \\
 Linear warmup (optimizer updates) & 0 & 2,000 \\
 Precision & FP32 & BF16 autocast \\
 Optimizer / weight decay & AdamW / 0.01 & AdamW / 0.01 \\
 Adam coefficients / epsilon & $(0.9,0.999)$ / $10^{-8}$ & $(0.9,0.999)$ / $10^{-8}$ \\
 Global gradient-norm clipping & 1.0 & 1.0 \\
 Dropout / learning-rate decay & 0 / none & 0 / none \\
 Seeds & 11, 23, 37 & 11, 23, 37 \\
 \bottomrule
 \end{tabular}
\end{table}
All models are trained with AdamW \cite{adam,adamw} using the
size-specific settings in Table~\ref{tab:recipe}. Each run performs
20,000 optimizer updates and processes 81.92M scored targets.
The small models use eight sequences per update in FP32. The larger
models use microbatches of four sequences, accumulate two microsteps
per update, and use BF16 autocast. Their learning rate increases
linearly to $10^{-4}$ over the first 2,000 updates and then remains
constant. Each run uses one A100 GPU and eight CPU cores.

Training blocks are shuffled once and sampled without replacement
until the fixed target budget is reached. The reported endpoint is
the saved final checkpoint after this schedule. The full matrix comprises
six architectures, three corpora, two sizes, and seeds 11, 23, and 37,
for a total of 108 training runs; the retention study adds nine runs.
Exploratory held-out results informed architecture development before
the complete matrix was evaluated.

Initialization and shuffled loading use a shared PyTorch random stream
in the reported experiments. Since architectures have different parameter
shapes, identical seed labels identify repeated runs but do not impose
identical minibatch orders across models. All runs nevertheless process
the same number of targets under the same size-specific schedule.
Appendix~\ref{app:repro} specifies this sampling protocol and the
independent-sampler option provided for subsequent experiments.

The principal metric is negative log-likelihood over the full held-out
evaluation stream and its exponential, perplexity. We denote this
loss by $\mathcal L_{\mathrm{eval}}$.
For $K$ scored held-out targets, negative log-likelihood (NLL) is
\begin{equation}
 \mathcal L_{\mathrm{eval}}=-\frac1K\sum_{n=1}^K
 \log P_\theta(y_n\mid x_{\mathrm{prefix}(n)}),\qquad
 \mathrm{PPL}=\exp(\mathcal L_{\mathrm{eval}}).
 \label{eq:metric}
\end{equation}
$n$ indexes all scored targets in complete blocks; $y_n$ is the
true symbol and $x_{\mathrm{prefix}(n)}$ its within-block input prefix.
Summed loss is divided by total targets, including the last short batch.
On raw enwik8/TinyStories, bits per scored byte equal
$\mathcal L_{\mathrm{eval}}/\log2$. WikiText also scores artificial
separators, so its metric is per symbol, not strictly per byte.
The resulting perplexity is measured per scored byte or separator symbol,
rather than per word token. All results in the following section use
the complete held-out stream at the final checkpoint. The mean and
sample standard deviation summarize the three run-level scores for
each condition; Appendix~\ref{app:seeds} reports the individual runs.

\FloatBarrier

\section{Results}
\label{sec:results}
\subsection{Held-out performance across corpora and sizes}
Tables~\ref{tab:small} and \ref{tab:medium} report final-checkpoint
perplexity for every architecture, corpus, size and seed, summarized
as a mean and sample standard deviation. The causal workspace has
the best mean in five of the six corpus--size comparisons. This
pattern is especially clear at 101.71M parameters: it leads all
three corpora against controls that use the same number of shared
block applications and nearly the same parameter count.
\begin{table}[t]
 \centering\small
 \caption{Small-budget held-out PPL, mean $\pm$ sample standard
 deviation across seeds 11, 23, 37. Lower is better; bold marks the
 lowest mean per corpus. Each run processes 81.92M scored targets.}
 \label{tab:small}\begin{tabular}{lrrr}
\toprule
Architecture & WikiText-103 & enwik8 & TinyStories \\
\midrule
Plain Transformer & $4.5884 \pm 0.0751$ & $4.7063 \pm 0.0558$ & $2.5923 \pm 0.0718$ \\
Fixed shared loop & $4.1668 \pm 0.1420$ & $4.4215 \pm 0.2663$ & $2.3389 \pm 0.0030$ \\
Adaptive-exit mixture & $4.1305 \pm 0.0216$ & $4.4937 \pm 0.2993$ & $2.4100 \pm 0.0631$ \\
Anchored shared loop & $4.0022 \pm 0.0201$ & $3.9125 \pm 0.0482$ & $2.2336 \pm 0.0738$ \\
Anchored conditioned-dense & $3.9481 \pm 0.0069$ & $\mathbf{3.8431 \pm 0.0510}$ & $2.2225 \pm 0.0533$ \\
Causal workspace (ours) & $\mathbf{3.7455 \pm 0.0042}$ & $3.8968 \pm 0.2019$ & $\mathbf{2.1732 \pm 0.0151}$ \\
\bottomrule
\end{tabular}

\end{table}
\begin{table}[t]
 \centering\small
 \caption{Larger-budget (101.71M) held-out PPL at the same
 scored-target budget. Means, variation and bolding follow
 Table~\ref{tab:small}. These are final-checkpoint scores.}
 \label{tab:medium}\begin{tabular}{lrrr}
\toprule
Architecture & WikiText-103 & enwik8 & TinyStories \\
\midrule
Plain Transformer & $3.0300 \pm 0.0519$ & $3.0178 \pm 0.0217$ & $1.8033 \pm 0.0171$ \\
Fixed shared loop & $2.8889 \pm 0.0077$ & $2.8810 \pm 0.0025$ & $1.7416 \pm 0.0017$ \\
Adaptive-exit mixture & $2.8950 \pm 0.0033$ & $2.8803 \pm 0.0014$ & $1.7395 \pm 0.0030$ \\
Anchored shared loop & $2.8118 \pm 0.0063$ & $2.8049 \pm 0.0081$ & $1.7040 \pm 0.0025$ \\
Anchored conditioned-dense & $2.8136 \pm 0.0075$ & $2.8057 \pm 0.0022$ & $1.6996 \pm 0.0032$ \\
Causal workspace (ours) & $\mathbf{2.7820 \pm 0.0032}$ & $\mathbf{2.7843 \pm 0.0010}$ & $\mathbf{1.6823 \pm 0.0015}$ \\
\bottomrule
\end{tabular}

\end{table}
At 0.918M parameters (Table~\ref{tab:small}), the workspace
reaches 3.7455 PPL on WikiText-103 and 2.1732 on TinyStories.
The strongest alternative on both corpora is the anchored
conditioned-dense loop, at 3.9481 and 2.2225; the workspace
reduces perplexity by 5.13\% and 2.22\%, respectively.
The WikiText advantage appears for all three seed labels, and
the TinyStories advantage for two. On small enwik8, the
conditioned-dense loop instead leads with 3.8431 versus the
workspace's 3.8968. The workspace's 0.2019 standard deviation
there is also larger than in its other small-model cells. Thus
the proposed structure is the leading small-model design on
two distinct corpora, but not every byte distribution.

At 101.71M parameters (Table~\ref{tab:medium}), the workspace
achieves 2.7820, 2.7843 and 1.6823 PPL on WikiText-103,
enwik8 and TinyStories. The strongest competing means are
2.8118 and 2.8049 from anchored looping on the first two
corpora, and 1.6996 from conditioned-dense looping on
TinyStories. Consequently, the workspace improves upon the
best non-workspace alternative by 1.06\%, 0.73\% and 1.01\%.
It leads for all three seed labels on each corpus, rather
than owing its ranking to one favorable run. On enwik8 its
mean NLL corresponds to 1.4773 bits per byte.

\begin{figure}[t]
 \centering\includegraphics[width=\linewidth]{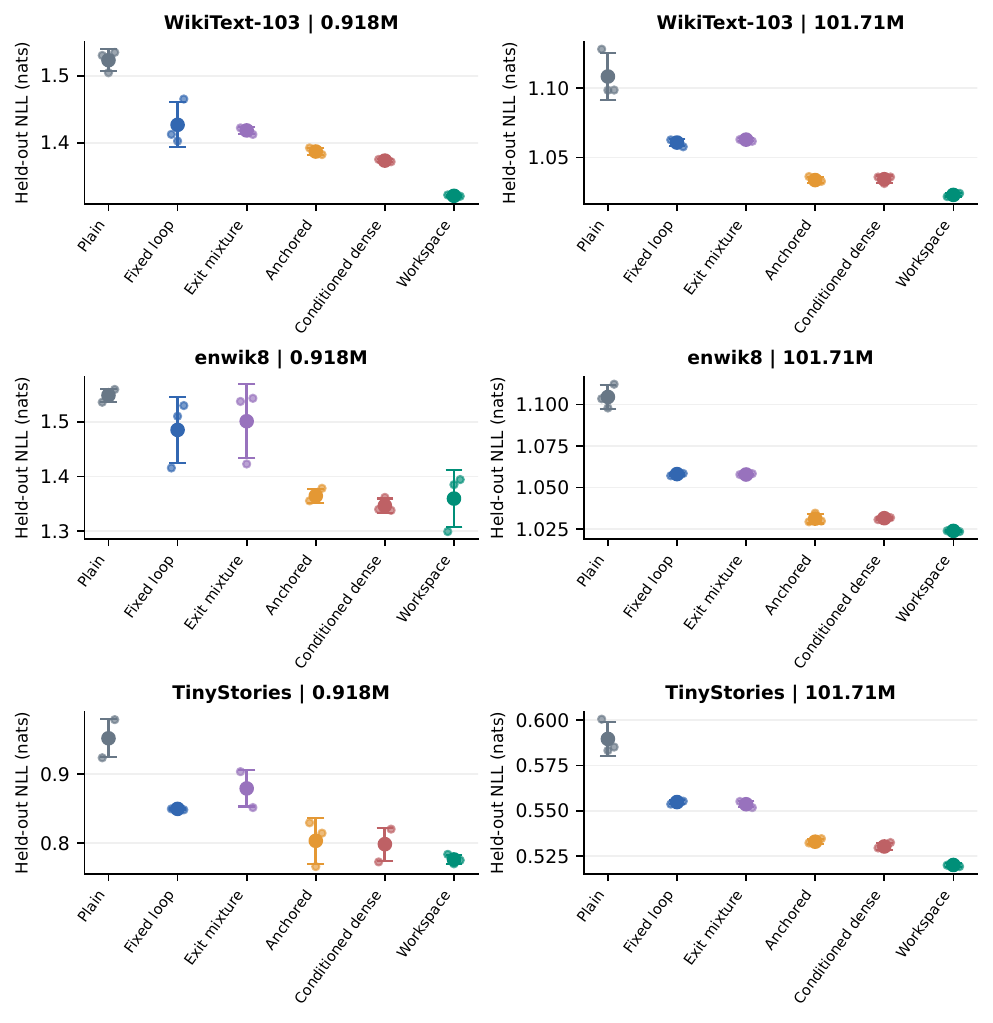}
 \caption{Complete-grid held-out NLL. Large markers/bars show
 three-run means/sample standard deviations; small offset markers
 show seeds. Rows are corpora, columns are parameter budgets.
 Each point represents the final checkpoint evaluated over the full held-out stream.}
 \label{fig:validation}
\end{figure}

The pattern across corpora and sizes is summarized in
Figure~\ref{fig:validation}, which displays the endpoint NLLs
and their variation across runs. Each row corresponds to a corpus
and each column to a parameter budget. The larger-model panels
show the workspace advantage on all three corpora, while the
small-enwik8 panel shows the stronger conditioned-dense mean
and the larger spread of the workspace runs. The plotted NLLs
and the perplexities in Tables~\ref{tab:small}--\ref{tab:medium}
express the same endpoint scores on different scales.

Against the plain Transformer, the workspace's mean perplexity
is lower by 16.17--18.37\% at the small budget and
6.71--8.19\% at the larger budget. This comparison establishes
the benefit of the complete architecture. The shared-loop,
anchored, and conditioned-dense rows then locate the gain more
precisely: weight sharing and anchoring both improve performance,
but neither consistently reaches the workspace result. In
particular, adding the conditioned-dense transition to the
anchored loop helps small models and larger TinyStories while
slightly worsening larger WikiText-103 and enwik8. Merely adding
more expressive recurrence is therefore not sufficient to explain
the workspace's three-corpus advantage at 101.71M.

\subsection{Architectural and retention ablations}
Table~\ref{tab:ablation} isolates two successive design decisions
at 101.71M parameters. The anchored-loop row \emph{removes the
causal workspace path}: it has no gated slot writes, learned-key
reads, or workspace residual, but keeps the input anchor and
four applications of the shared causal block. To retain the
parameter budget, its feed-forward sublayer is wider. Restoring
the workspace produces the second row, reducing mean PPL from
2.8118 to 2.7820 on WikiText-103, 2.8049 to 2.7843 on enwik8,
and 1.7040 to 1.6823 on TinyStories. Thus the workspace improves
prediction on every larger-budget corpus even when the anchored
alternative reallocates its parameters to a wider feed-forward
sublayer. The matched-budget comparison supports the complete causal
write/read pathway as an effective architectural addition.
\begin{table}[t]
 \centering\small
 \caption{Workspace-path and across-round retention ablations at 101.71M. Anchored and
 recomputed-workspace rows are from the grid; retention adds nine
 runs at the same training budget. Held-out PPL,
 mean $\pm$ sample standard deviation.}
 \label{tab:ablation}\begin{tabular}{lrrr}
\toprule
Architecture & WikiText-103 & enwik8 & TinyStories \\
\midrule
Anchored shared loop & $2.8118 \pm 0.0063$ & $2.8049 \pm 0.0081$ & $1.7040 \pm 0.0025$ \\
Causal workspace (ours) & $2.7820 \pm 0.0032$ & $2.7843 \pm 0.0010$ & $1.6823 \pm 0.0015$ \\
Workspace + loop retention & $2.7794 \pm 0.0092$ & $2.7863 \pm 0.0082$ & $1.6827 \pm 0.0020$ \\
\bottomrule
\end{tabular}

\end{table}
The third row changes only \emph{how slot state persists between
rounds}. The main workspace model recomputes each prefix summary
from the current round's token states. The retention variant
replaces it with the learned convex blend in
Equation~\ref{eq:retention}, while preserving the same shared
block, write/read pathway and nominal parameter count.
Its mean PPLs are 2.7794, 2.7863 and 1.6827. Both workspace
variants outperform the anchored loop on all three corpora, supporting
the value of the causal summary pathway under either update rule.
The choice of update rule has a much smaller effect: relative to
recomputation, retention changes mean PPL by $-0.0931\%$ on WikiText-103,
$+0.0722\%$ on enwik8, and $+0.0216\%$ on TinyStories.
Figure~\ref{fig:ablation-cost}A resolves these changes into individual
seed-labelled NLL differences. The main model uses recomputation,
which achieves the lower mean on two corpora and refreshes each
prefix summary using the current token states. The architectural comparison thus supports the added causal
summary pathway, while the retention comparison determines how
that pathway is updated across depth. The roles of gating, learned
keys, and the read projection can be examined separately through
further component ablations.

\subsection{Training behavior and computational cost}
\begin{figure}[t]
 \centering\includegraphics[width=\linewidth]{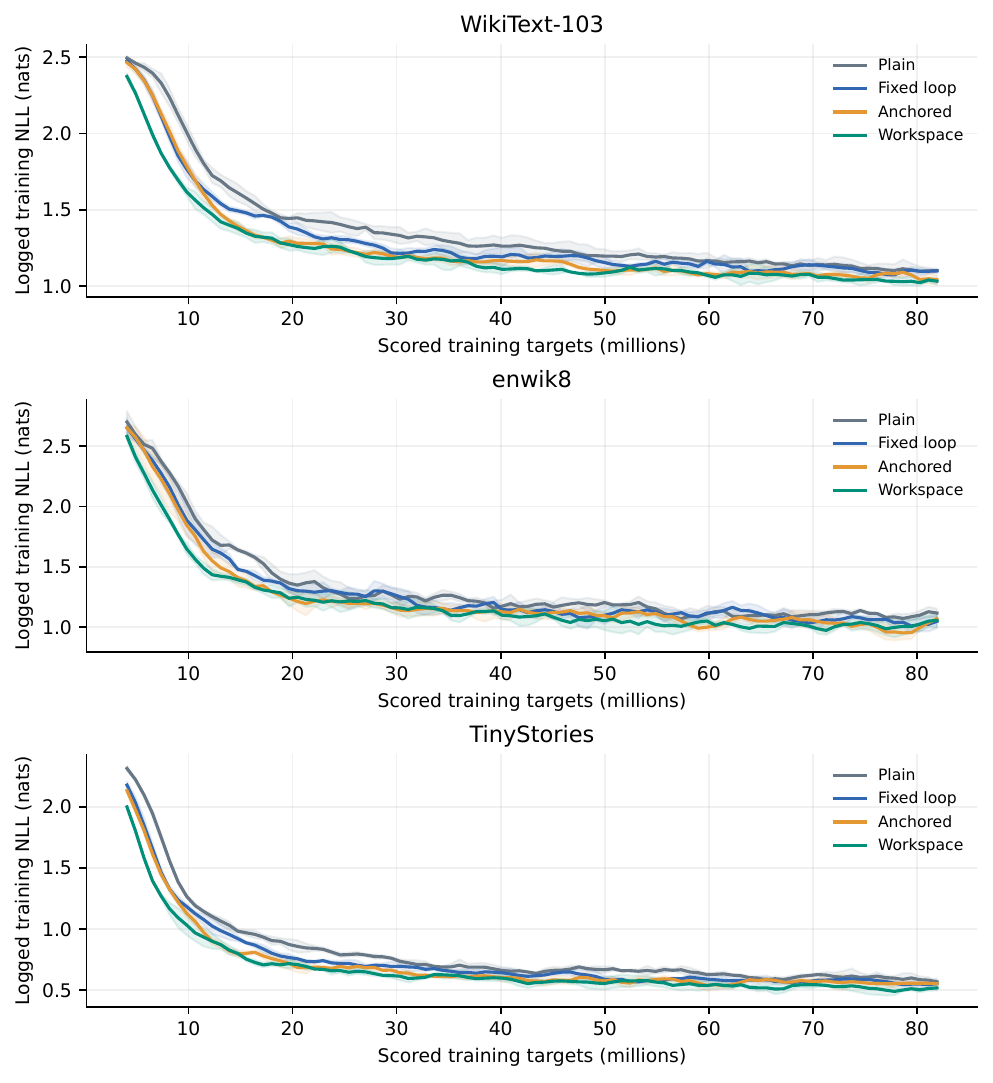}
 \caption{101.71M training NLL for four representative rows. Each curve
 is a five-observation moving average of periodically logged
 microbatch loss, then averaged over seeds; shading shows sample
 standard deviation of smoothed seed curves.
 Each run contributes 101 periodically logged training observations.}
 \label{fig:training}
\end{figure}

The endpoint results can be considered alongside the course of
optimization. Figure~\ref{fig:training} shows smoothed training NLL
for four representative 101.71M architectures over the common
81.92M-target schedule. The shared-depth models generally reach
lower training loss than the plain Transformer, and the workspace
follows the favorable recurrent-model trajectory. These curves
describe the periodically sampled training minibatches, complementing
the held-out endpoint comparisons in Table~\ref{tab:medium}.
\begin{figure}[t]
 \centering\includegraphics[width=\linewidth]{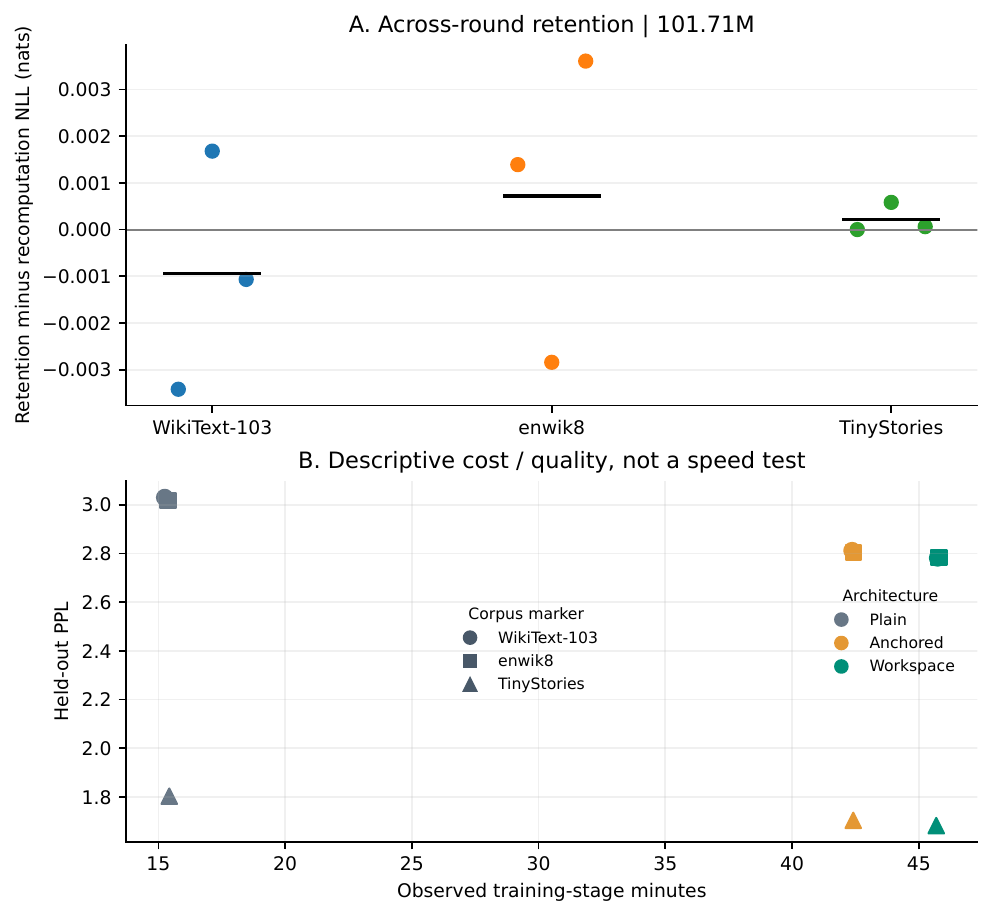}
 \caption{A: retention minus recomputed-workspace held-out NLL;
 negative favors retention. Points are seed-labelled differences;
 black lines mark means. B: observed training-stage minutes versus
 held-out PPL. Color identifies architecture; circle/square/triangle
 denote WikiText/enwik8/TinyStories.
 Training-stage timing includes checkpoint I/O and a limited held-out pass;
 it excludes setup and the complete held-out evaluation.}
 \label{fig:ablation-cost}
\end{figure}
The additional projections and prefix accumulations also affect
training cost. Figure~\ref{fig:ablation-cost}B relates the observed
training-stage time to held-out perplexity for the recorded 101.71M
runs. The workspace reaches the lowest mean perplexity on each corpus,
while its training-stage records reflect the work of the added
write/read pathway. Timings include checkpoint I/O and the limited
held-out pass performed by the trainer, so they describe the observed
runs rather than a controlled kernel-throughput measurement.
The resulting quality--cost comparison characterizes the additional
computation associated with improved prediction at the matched budgets.

\subsection{Contribution of recurrent depth and the workspace}
The anchored shared loop provides the most direct architectural
comparison because it preserves the input anchor and executes the same
four shared rounds. Adding the workspace lowers its mean perplexity in
all six corpus--size settings. At 0.918M parameters, the reductions on
WikiText-103, enwik8, and TinyStories are 6.41\%, 0.40\%, and 2.70\%;
at 101.71M they are 1.06\%, 0.73\%, and 1.27\%, respectively.
Thus the workspace improves upon anchoring even in the small-enwik8
setting where the conditioned-dense architecture has the best overall
mean. This comparison identifies a consistent benefit from allocating
part of the matched budget to the causal summary pathway.

The other controls establish how this benefit relates to recurrent
depth. Applying a shared block four times improves upon the plain
Transformer in all six settings, and adding the input anchor improves
upon the fixed shared loop on every corpus at both sizes. The workspace
builds on these improvements by introducing content-addressed prefix
summaries. The conditioned-dense control instead modulates the recurrent
transition and combines earlier exits. Its ranking relative to anchoring
varies across corpora, whereas the workspace improves upon the anchored
loop throughout the grid and leads the complete larger-model comparison.

The repeated larger-model ranking across WikiText-103, enwik8, and
TinyStories indicates that the benefit extends across the three tested
text distributions. Absolute perplexities reflect each corpus's byte
distribution and are therefore interpreted within corpus. Together with
the retention results, these comparisons support the architectural
combination of anchored recurrence and a refreshed causal workspace:
the anchor preserves the original input, and the workspace supplies a
distinct summary of information accumulated over the available prefix.

\FloatBarrier

\section{Conclusion and future work}
We introduced a causal latent workspace Transformer that combines
anchored shared-depth refinement with gated causal writes and learned-key
reads. The workspace makes compact, prefix-specific summaries available
before each shared block application, providing an additional information
pathway alongside token-level self-attention. This design improves upon
the anchored loop in all six tested corpus--size settings and achieves
the lowest mean held-out perplexity among six architectures in five.

At 101.71M parameters, the workspace leads on WikiText-103 raw, enwik8,
and TinyStories, with all corresponding seed-labelled comparisons
favoring it over the strongest non-workspace control. At 0.918M
parameters it leads on WikiText-103 and TinyStories; conditioned-dense
looping is strongest on enwik8. The retention study further shows that
both tested workspace update rules improve upon the larger anchored
control on every corpus, with direct recomputation giving the lower mean
on two of the three. These findings support a causal write/read pathway
as an effective structural addition to looped Transformers at the
evaluated parameter and training-target budgets.

The next architectural question concerns the individual contributions
of gating, addressing, and residual injection. Experiments that hold
hidden and feed-forward widths fixed while varying these operations can
clarify how they contribute to the complete pathway. Longer training
schedules can assess the persistence of the observed advantage, while
longer contexts and subword language modeling can test the mechanism
under different sequence representations. Incremental decoding and
measurements of activation memory and compute-matched quality will
further characterize the trade-offs of workspace-assisted refinement.

\section*{Code, data and reproducibility}
The implementation, experiment configurations, per-seed results, and
sources for the figures and tables are available at
\begin{center}
\small\url{https://github.com/izaznov/causal-latent-workspace-transformer}
\end{center}
The repository includes the six benchmark architectures, the retention
variant, and the offline training and evaluation procedures.
Dataset sources are linked in Section~\ref{sec:protocol}.
Appendix~\ref{app:repro} specifies the implementation mapping,
environment requirements, and commands for reproducing the study.

\clearpage
\appendix

\section{Symbol definitions and shared-block equations}
\label{app:block}
\begin{table}[h]
 \centering\small
 \caption{Main notation. Learnable weights, biases and LayerNorm
 parameters are shared over rounds unless round-indexed.}
 \begin{tabular}{ll}
 \toprule
 Symbol & Meaning \\
 \midrule
 $\mathcal V,x_i,y_i$ & Vocabulary, input symbol, next-symbol target \\
 $L,R,S,m,d,f,H$ & Context, rounds, slots, slot width, state width, FFN width, heads \\
 $B,K$ & Microbatch sequences and scored held-out-target count \\
 $e_i,h_i^r,u_i^r,z_i^r$ & Anchor, recurrent state, fused state, write-normalized state \\
 $g_{is}^r,v_{is}^r$ & Scalar write gate and value vector \\
 $N_{is}^r,D_{is}^r$ & Gated prefix sum and scalar gate mass \\
 $\widetilde M_{is}^r,M_{is}^r$ & Newly written and actually read memory \\
 $\ell_r,\rho_r$ & Retention logit and sigmoid \\
 $q_i^r,k_s,\alpha_{is}^r,c_i^r$ & Query, slot key, read weight, residual \\
 $\gamma,\epsilon$ & Workspace gain and prefix-mass numerical floor \\
 $T_\theta,\theta$ & Shared block and all slow parameters \\
 $o_i,W_{\mathrm{LM}},\mathcal L_{\mathrm{eval}}$ & Logits, head matrix, held-out NLL \\
 \bottomrule
 \end{tabular}
\end{table}
Let $t_i\in\R^d$ denote one block's input, and let $H$ be the head
count with width $d_h=d/H$. For head $a=1,\ldots,H$, normalized input
$\bar t_i=\LN_1(t_i)$ produces $Q_i^a,K_i^a,V_i^a\in\R^{d_h}$
by learned affine projections. Capital $K_i^a$ is an attention key,
distinct from the held-out target count $K$ and workspace key $k_s$.
Attention probabilities and head outputs are
\begin{equation}
 A_{ij}^a=
 \begin{cases}
 \displaystyle\frac{\exp((Q_i^a)^\top K_j^a/\sqrt{d_h})}
 {\sum_{k=1}^{i}\exp((Q_i^a)^\top K_k^a/\sqrt{d_h})},&j\leq i,\\
 0,&j>i,
 \end{cases}\qquad
 b_i^a=\sum_{j=1}^{i}A_{ij}^a V_j^a .
\end{equation}
The dummy $k$ is a source position. $A_{ij}^a$ is a probability;
$b_i^a$ is a head output, not an affine bias. With learned output
matrix $W_A\in\R^{d\times d}$ and bias $b_A\in\R^d$,
\begin{align}
 y_i^{\mathrm{blk}}&=t_i+W_A[b_i^1;\ldots;b_i^H]+b_A,\\
 T_\theta(t)_i&=y_i^{\mathrm{blk}}+
 W_2\,\mathrm{GELU}(W_1\LN_2(y_i^{\mathrm{blk}})+b_1)+b_2 .
\end{align}
$y_i^{\mathrm{blk}}$ is a block intermediate, not target $y_i$.
$W_1\in\R^{f\times d}$, $b_1\in\R^f$,
$W_2\in\R^{d\times f}$ and $b_2\in\R^d$ are learned.
All dropout is zero. Each LayerNorm on vector $z\in\R^d$ uses its
own learned gain $\lambda\in\R^d$ and offset $\delta\in\R^d$:
\begin{equation}
 \LN(z)=\lambda\odot\frac{z-\mu(z)\mathbf1}{\sqrt{\nu(z)+\eta}}+\delta,
 \quad \mu(z)=\frac1d\sum_{a=1}^d z_a,\quad
 \nu(z)=\frac1d\sum_{a=1}^d(z_a-\mu(z))^2,\quad \eta=10^{-5}.
\end{equation}
$\odot$ denotes coordinate-wise multiplication, $\mathbf1$ the
all-ones vector, $\mu$ the coordinate mean, $\nu$ population coordinate
variance and $\eta$ LayerNorm epsilon. Here $a$ indexes coordinates,
not heads. Gain and offset initialize to one and zero.

\section{Exact architecture dimensions}
\label{app:hyper}
\begin{table}[h]
 \centering\small
 \caption{Exact dimensions and nominal trainable counts. FFN width is $f$.
 Dash denotes no workspace. Untied blocks / rounds are 4 / 1 for plain
 and 1 / 4 for recurrent rows.}
 \begin{tabular}{llrrrrr}
 \toprule
 Budget & Architecture & $d$ & $H$ & $f$ & $S\times m$ & Parameters \\
 \midrule
 Small & Plain & 128 & 4 & 506 & -- & 918,504 \\
 & Fixed loop & 256 & 4 & 761 & -- & 918,009 \\
 & Exit mixture & 256 & 4 & 759 & -- & 918,011 \\
 & Anchored & 215 & 5 & 860 & -- & 918,050 \\
 & Conditioned dense & 215 & 5 & 856 & -- & 918,049 \\
 & Workspace & 215 & 5 & 758 & $4\times32$ & 918,030 \\
 \midrule
 Larger & Plain & 1,536 & 12 & 5,071 & -- & 101,715,772 \\
 & Fixed loop & 3,072 & 24 & 9,890 & -- & 101,708,450 \\
 & Exit mixture & 3,072 & 24 & 9,888 & -- & 101,708,452 \\
 & Anchored & 2,560 & 20 & 10,383 & -- & 101,711,503 \\
 & Conditioned dense & 2,560 & 20 & 10,379 & -- & 101,711,502 \\
 & Workspace & 2,560 & 20 & 9,993 & $4\times128$ & 101,709,714 \\
 \bottomrule
 \end{tabular}
\end{table}
The exit mixture uses
$\pi_i=\softmax(W_{\mathrm{exit}}e_i+b_{\mathrm{exit}})$
and final state $\sum_{r=0}^{R-1}\pi_{i,r}h_i^{r+1}$.
$W_{\mathrm{exit}}\in\R^{R\times d}$,
$b_{\mathrm{exit}}\in\R^R$ are learned; $\pi_{i,r}$ weights exit $r+1$.
The conditioned-dense control sends
$(1+\sigma_r)\odot u_i^r+\tau_r$ to the block, with learned
$\sigma_r,\tau_r\in\R^d$ initialized to zero. Its readout is
\begin{equation}
 \widehat h_i=h_i^R+\sum_{r=0}^{R-2}\tanh(\beta_r)(h_i^{r+1}-h_i^R).
\end{equation}
$\beta_r$ are zero-initialized learned scalars.
$\widehat h_i$ replaces $h_i^R$ before the final normalization/head.
These modifications change recurrent transition and readout,
not the number of independently parameterized blocks.

Parameter counts include position embeddings and the vocabulary head.
Both workspace variants allocate $R$ scalar retention logits to keep
their nominal counts identical. The recomputed variant does not use
these logits; in the retained variant, $\ell_0$ is unused because the
first round has no preceding workspace. Projection matrices use the
standard PyTorch initializers, with the slot-key, gain, retention,
and conditioned-dense initializations specified above.

\section{Complete per-seed scores}
\label{app:seeds}
\begin{table}[h]
 \centering\small
 \caption{All small-budget full held-out PPLs. The mean is the arithmetic
 mean of run-level PPLs, not exponentiated mean NLL.}
 \begin{tabular}{llrrrr}
\toprule
Dataset & Architecture & Seed 11 & Seed 23 & Seed 37 & Mean \\
\midrule
WikiText-103 & Plain & 4.62059 & 4.50256 & 4.64196 & 4.58837 \\
 & Fixed loop & 4.10625 & 4.06510 & 4.32895 & 4.16677 \\
 & Exit mixture & 4.14598 & 4.13962 & 4.10585 & 4.13048 \\
 & Anchored & 4.02385 & 3.99855 & 3.98409 & 4.00216 \\
 & Conditioned dense & 3.95518 & 3.94761 & 3.94147 & 3.94809 \\
 & Workspace & 3.75042 & 3.74290 & 3.74327 & 3.74553 \\
\midrule
enwik8 & Plain & 4.64615 & 4.71639 & 4.75645 & 4.70633 \\
 & Fixed loop & 4.11837 & 4.52836 & 4.61779 & 4.42151 \\
 & Exit mixture & 4.65297 & 4.14851 & 4.67974 & 4.49374 \\
 & Anchored & 3.87723 & 3.89291 & 3.96748 & 3.91254 \\
 & Conditioned dense & 3.81711 & 3.90177 & 3.81028 & 3.84305 \\
 & Workspace & 3.66466 & 3.99426 & 4.03161 & 3.89684 \\
\midrule
TinyStories & Plain & 2.51923 & 2.59495 & 2.66279 & 2.59232 \\
 & Fixed loop & 2.33897 & 2.34188 & 2.33582 & 2.33889 \\
 & Exit mixture & 2.46892 & 2.41779 & 2.34344 & 2.41005 \\
 & Anchored & 2.29224 & 2.15082 & 2.25788 & 2.23364 \\
 & Conditioned dense & 2.16544 & 2.23084 & 2.27111 & 2.22246 \\
 & Workspace & 2.18941 & 2.15944 & 2.17081 & 2.17322 \\
\bottomrule
\end{tabular}

\end{table}
\begin{table}[h]
 \centering\small
 \caption{All larger-budget full held-out PPLs. Matching seed labels
 do not establish identical minibatch order across rows.}
 \begin{tabular}{llrrrr}
\toprule
Dataset & Architecture & Seed 11 & Seed 23 & Seed 37 & Mean \\
\midrule
WikiText-103 & Plain & 3.08994 & 2.99986 & 3.00033 & 3.03004 \\
 & Fixed loop & 2.89454 & 2.89208 & 2.88015 & 2.88893 \\
 & Exit mixture & 2.89524 & 2.89818 & 2.89159 & 2.89500 \\
 & Anchored & 2.81900 & 2.80726 & 2.80927 & 2.81184 \\
 & Conditioned dense & 2.81796 & 2.80494 & 2.81803 & 2.81364 \\
 & Workspace & 2.77861 & 2.78226 & 2.78506 & 2.78198 \\
\midrule
enwik8 & Plain & 3.01451 & 2.99787 & 3.04099 & 3.01779 \\
 & Fixed loop & 2.87815 & 2.88271 & 2.88212 & 2.88099 \\
 & Exit mixture & 2.88017 & 2.87891 & 2.88179 & 2.88029 \\
 & Anchored & 2.79959 & 2.81425 & 2.80082 & 2.80488 \\
 & Conditioned dense & 2.80318 & 2.80728 & 2.80652 & 2.80566 \\
 & Workspace & 2.78448 & 2.78520 & 2.78316 & 2.78428 \\
\midrule
TinyStories & Plain & 1.82290 & 1.79173 & 1.79515 & 1.80326 \\
 & Fixed loop & 1.73964 & 1.74273 & 1.74245 & 1.74161 \\
 & Exit mixture & 1.74217 & 1.74012 & 1.73626 & 1.73952 \\
 & Anchored & 1.70284 & 1.70233 & 1.70690 & 1.70402 \\
 & Conditioned dense & 1.69816 & 1.69730 & 1.70320 & 1.69955 \\
 & Workspace & 1.68210 & 1.68389 & 1.68095 & 1.68231 \\
\bottomrule
\end{tabular}

\end{table}
\begin{table}[h]
 \centering\small
 \caption{Retention PPL by seed. Last column: change in mean PPL from
 recomputed workspace; negative is better.}
 \begin{tabular}{lrrrr}
\toprule
Dataset & Seed 11 & Seed 23 & Seed 37 & Mean PPL change (\%) \\
\midrule
WikiText-103 & 2.76913 & 2.78694 & 2.78209 & -0.0931 \\
enwik8 & 2.78836 & 2.77730 & 2.79322 & +0.0722 \\
TinyStories & 1.68210 & 1.68488 & 1.68106 & +0.0216 \\
\bottomrule
\end{tabular}

\end{table}
Full-precision NLL, PPL, raw-byte bits-per-byte values, parameters,
evaluation-block counts, hashes, training times and run IDs for all 117
evaluations are in \code{paper/manuscript/results.csv}.
Source records and 108 logged trajectories are in
\code{paper/manuscript/evidence.json}.
\FloatBarrier

\section{Fully specified numerical toy}
\label{app:toy}
The following parameter choices specify the numerical example in
Section~\ref{sec:walkthrough}.
Set $d=S=m=2$, $R=4$ and $E(x_i)+p_i=(0.3,-0.1)$ for all eight
positions; use identical embeddings and zero position vectors.
Set all $F_a$ parameters to zero and LayerNorm gain/offset to
one/zero with $\eta=10^{-5}$.
Set $W_g=0$, $b_g=(\log(1/3),0)$, giving gates $(0.25,0.5)$.
Set $W_v=0$ and
\[
 b_v=(\operatorname{atanh}(0.4),\operatorname{atanh}(0.2),
       \operatorname{atanh}(0.2),\operatorname{atanh}(-0.4)).
\]
$\operatorname{atanh}$ is inverse hyperbolic tangent.
Choose $W_q=0$, $b_q=(1,0)$, $k_1=(1,0)$, $k_2=(0,1)$,
$W_o=I_2$, $b_o=0$, and $\gamma=0.5$, where $I_2$ is the
two-dimensional identity. All attention/FFN projection weights and
biases are zero: their residual sublayers return the input, so
$T_\theta$ is identity. Recompute the workspace each round.

At position eight, the exact slot-one weight and workspace residual are
\[
 a_*=\frac{\exp(1/\sqrt2)}{\exp(1/\sqrt2)+1},\qquad
 c_*=\tfrac12\left(a_*(0.8,0.4)/\sqrt2+(1-a_*)(0.4,-0.8)\right).
\]
$a_*$ is a scalar and $c_*$ a two-coordinate residual.
The recurrence is $u^r=\LN(h^r)$, $h^{r+1}=\LN(u^r+c_*)$,
starting at $h^0=(0.3,-0.1)$. It gives
$u^0\approx(0.999875,-0.999875)$,
$h^1\approx(0.999996,-0.999996)$; subsequent rounds stay nearby.
The resulting intermediate values are tabulated for all four rounds
in the accompanying numerical summary.

Write $\LN_f(h^4)=(c_f,-c_f)$, with positive scalar $c_f$ computed
using the same $\eta$. Choose output rows
$(3/(2c_f),-3/(2c_f))$ for \texttt{s},
$(1/(2c_f),-1/(2c_f))$ for \texttt{a}, $(0,0)$ for space,
and $(-5/(2c_f),5/(2c_f))$ for each other class.
This construction gives the logits used in the walkthrough through
the same linear head and softmax as Equation~\ref{eq:head}.
Identical embeddings and zero attention weights make the arithmetic
explicit. In the trained models, byte-dependent embeddings and
learned attention supply the contextual representations from which
workspace writes and reads are constructed.

\section{Reproduction and provenance}
\label{app:repro}
\paragraph{Implementation map.}
\path{baselines/workspace_loop.py} implements the workspace;
\path{baselines/looped_transformer.py} recurrent controls;
\path{baselines/causal_transformer.py} the plain model.
\path{data/wikitext.py}, \path{data/enwik8.py} and
\path{data/raw_bytes.py} construct blocks.
\code{train_lm.py} trains and \code{evaluate_lm.py} evaluates
all held-out batches by default.
The matrix script is
\path{hpc/gwdg/standard_transformer_benchmark_matrix.sbatch},
with per-cell script \path{hpc/gwdg/standard_transformer_benchmark.sbatch}.
They record requested resources, offline paths, sequential seeds
and final-checkpoint evaluation.

\paragraph{Configurations and sampling.}
The small and larger recipes are specified by
\path{configs/standard_benchmark_small.yaml} and
\path{configs/standard_benchmark_medium.yaml}, respectively; retention
uses \path{configs/long_workspace_medium.yaml}.
The reported experiments use \code{--sampler-mode historical}, which
shares the PyTorch random stream between initialization and shuffled
loading. The release also provides
\code{--sampler-mode independent --data-seed 11} for experiments that
hold the data order fixed across architectures. Reproducing the
reported protocol requires the former setting. Exact bitwise replay
additionally depends on the original hardware and package versions;
executed revisions were not captured separately for each run.

Commands below illustrate one larger-budget WikiText workspace run, from the
repository root using downloaded official shards:
\begin{verbatim}
python train_lm.py --config configs/standard_benchmark_medium.yaml \
  --model anchored_workspace_static --seed 11 \
  --train-parquet DATA/wikitext-103-raw-train-0000.parquet \
    DATA/wikitext-103-raw-train-0001.parquet \
  --heldout-parquet DATA/wikitext-103-raw-heldout.parquet \
  --output-dir runs/replay-wiki-medium-workspace-11

python evaluate_lm.py --config configs/standard_benchmark_medium.yaml \
  --model anchored_workspace_static \
  --checkpoint runs/replay-wiki-medium-workspace-11/checkpoint.pt \
  --heldout-parquet DATA/wikitext-103-raw-heldout.parquet \
  --eval-batches 0 --output runs/replay-wiki-medium-workspace-11/full.json
\end{verbatim}
\noindent
\code{DATA} is a placeholder for the local data directory; replace it.
For enwik8, substitute \code{--enwik8-path DATA/enwik8} for
Parquet arguments. TinyStories uses \code{--raw-byte-train},
\code{--raw-byte-heldout} and \code{--max-train-tokens 250000000}
in training, and only \code{--raw-byte-heldout} in evaluation.
Repeat seeds 11, 23, 37, substitute architecture IDs, and use
the small configuration for the other budget.

\paragraph{Environment and auditing.}
Code requires Python 3.10+, PyTorch, NumPy, PyYAML, and PyArrow
for Parquet; matplotlib only rebuilds figures.
Historical scripts load Miniforge/GCC and activate a local environment.
The archive records SHA-256 hashes of each source
evaluation output and \code{metrics.json}, plus matched
main-grid log hashes. These identify the archived result records;
original dataset checksums and complete package-version locks were
not recorded. Independent replication should record these alongside
the sampler state and executed code revision.

\paragraph{Regenerate evidence displays.}
\code{python paper/manuscript/build_artifacts.py} verifies the
108 main and nine retention records, seed/cell coverage, target
budgets, parameter counts, NLL--PPL consistency and training-log
endpoint matches; it regenerates tables, CSV and vector figures.
The displayed curves use the archived training logs and complete
held-out scores. \code{tests/test_paper_models.py} verifies model
dimensions, causal dependence, and the workspace update equations.
\end{document}